\documentclass[10pt,twocolumn,letterpaper]{article}

\usepackage{cvpr}              % To produce the CAMERA-READY version

\usepackage[accsupp]{axessibility}

\usepackage{graphicx}
\usepackage{amsmath}
\usepackage{amssymb}
\usepackage{booktabs}
\usepackage{pifont}% http://ctan.org/pkg/pifont
\newcommand{\cmark}{\ding{51}}%
\newcommand{\xmark}{\ding{55}}%
\usepackage{caption}
\usepackage{subcaption}
\usepackage{adjustbox}

\let\subcaption\relax

\usepackage[pagebackref,breaklinks,colorlinks]{hyperref}

\usepackage[capitalize]{cleveref}
\crefname{section}{Sec.}{Secs.}
\Crefname{section}{Section}{Sections}
\Crefname{table}{Table}{Tables}
\crefname{table}{Tab.}{Tabs.}

\def\cvprPaperID{34} % *** Enter the CVPR Paper ID here
\def\confName{CVPR}
\def\confYear{2023}

\begin{document}

%\title{A deep metric learning approach for action segmentation}
\title{Leveraging triplet loss for unsupervised action segmentation}
\author{Elena Bueno-Benito  $^{\star}$ \\
{\tt\small ebueno@iri.upc.edu}
\and
Biel Tura Vecino\thanks{Work done during an internship at the IRI.} $^{\,\dagger}$ \\ %\thanks{Work done during an internship at the IRI.}biel.turavecino@gmail.com
{\tt\small  bieltura@amazon.co.uk}
\and
Mariella Dimiccoli $^{\star}$ \\
{\tt\small mdimiccoli@iri.upc.edu}
\and
$^{\star}$ Institut de Robòtica i Informàtica Industrial, CSIC-UPC, Barcelona, Spain \\
%$^{\dagger}$ Universitat Politècnica de Catalunya, UPC, Barcelona, Spain \\
$^{\dagger}$ Amazon Alexa AI, UK, Cambridge\\
}
 
\maketitle

%%%%%%%%% ABSTRACT
\begin{abstract}
   In this paper, we propose a novel fully unsupervised framework that learns action representations suitable for the action segmentation task from the single input video itself, without requiring any training data. 
   Our method is a deep metric learning approach rooted in a shallow network with a triplet loss operating on similarity distributions and a novel triplet selection strategy that effectively models temporal and semantic priors to discover actions in the new representational space. Under these circumstances, we successfully recover temporal boundaries in the learned action representations with higher quality compared with existing unsupervised approaches.
   The proposed method is evaluated on two widely used benchmark datasets for the action segmentation task and it achieves competitive performance by applying a generic clustering algorithm on the learned representations. \footnote{\url{https://github.com/elenabbbuenob/TSA-ActionSeg}}
   \vspace{-0.8em}
\end{abstract}

%%%%%%%%% BODY TEXT
\vspace{-1.0em}
\section{Introduction}
\label{sec:intro}
\vspace{-0.3em}
Unconstrained videos capturing real-world scenarios are usually long, untrimmed and contain a variety of actions which can be effortlessly divided by a human observer into semantically homogeneous units.
The task of action segmentation, which we target in this work, is the process of identifying the boundaries of an action, i.e. \textit{pour water}, in an untrimmed video of an activity, i.e. \textit{making tea}, even when temporally adjacent actions may have very small visual variance between them.  This process is a key step in understanding and contextualizing the video. It is also crucial for video browsing, indexing, and summarisation and has applications in areas such as surveillance systems, action recognition, video content-based retrieval, assistive technologies, and robot-human interactions.
\begin{figure}[t]
    \centering   
    \includegraphics[width=\columnwidth]{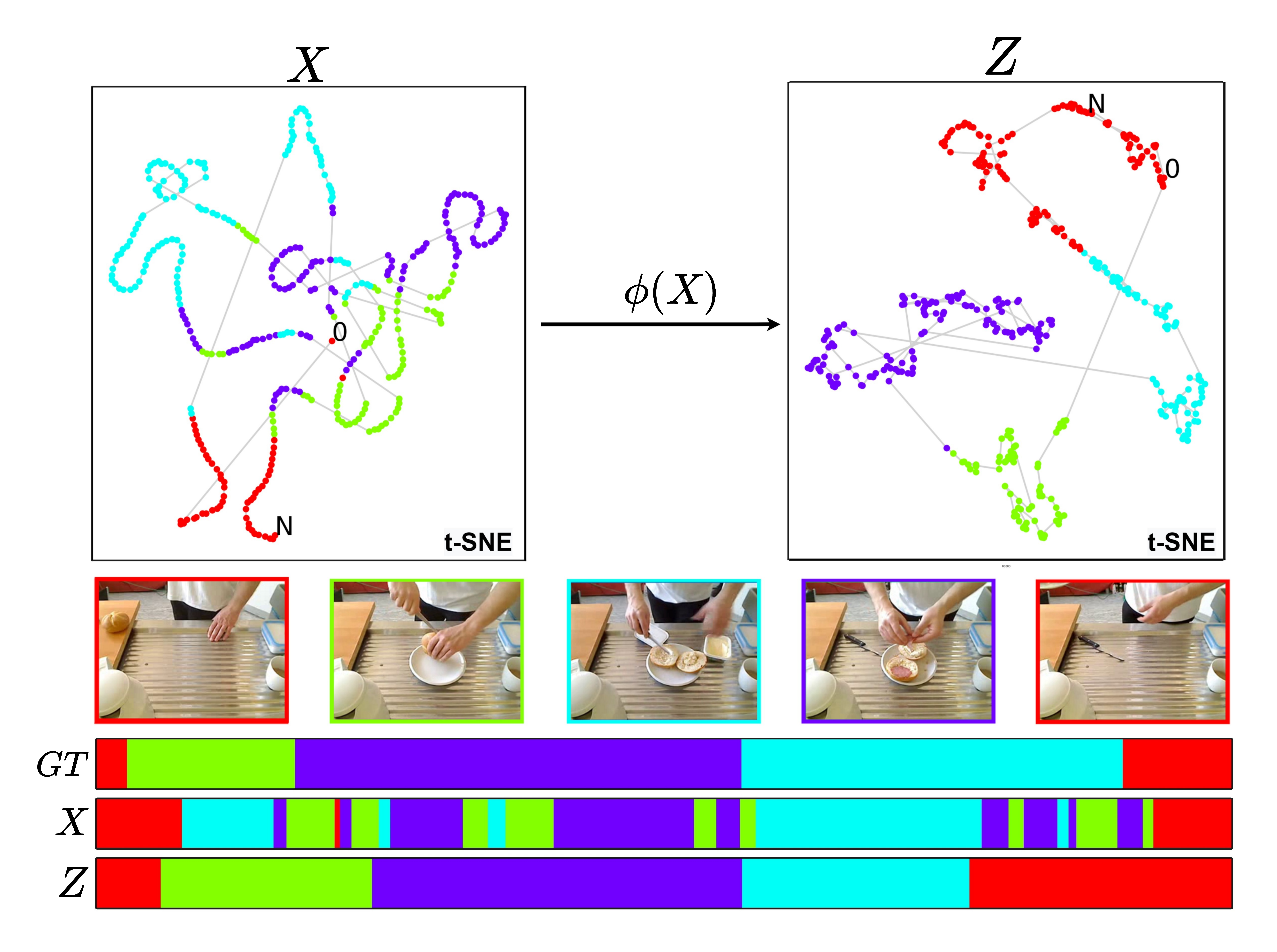}    
    \vspace{-1.7em}
    \caption{\footnotesize Our approach learns a parametrized function $\phi$ that transforms the input feature space ($X$) into a new one ($Z$), where actions, visualized through different colours, can be easily unveiled by a generic clustering algorithm. The continuous line connects points representing frames from time $0$ to $N$ in a t-SNE projection. GT stands for ground truth.}
    \vspace{-1.2em}
    \label{fig:tsne}
\end{figure}

This problem has been traditionally tackled through supervised learning approaches \cite{KuehneGS16, NgHVVMT15, LeaFVRH17, FarhaG19, li2020ms, LeaRVH16}. Such approaches require a large amount of annotated training data and typically suffer from domain adaptation problems, being unable to generalize to large-scale target domains. More recently, weakly-supervised and semi-supervised approaches have shown to be an effective way to learn video representations convenient for action segmentation without requiring or requiring very little manual annotations \cite{FayyazG20, KuehneRG17, RichardKG17, Lu_2021_ICCV, HuangFN16, NG2021103298, DingX18, RichardKIG18, LiLT19-8, AlayracBASLL16, SinghaniaRY22}. However, these approaches are still data-hungry and computationally expensive. Unsupervised approaches have developed following two different research lines \cite{Li2021ActionSA, Kumar_2022_CVPR, AakurS19}. Most of them focus on grouping actions across videos and rely on the use of activity labels \cite{KuklevaKSG19, SenerY18, VidalMataSKCK21, WangCLLXTF22, Li2021ActionSA, Kumar_2022_CVPR}, therefore putting more emphasis on the quality of the representation. A few of them, the most computationally efficient, act on a single video to recover clusters \cite{SarfrazM0DGS21} or detect temporal boundaries \cite{9879786} and do not require any manual annotation.  

Our approach stands between these two research lines. We assume that the atomic actions can effectively be modelled as clusters in an underlying representational space and we propose a novel framework that maps the initial feature space of a video into a new one, where the temporal-semantic clusters corresponding to atomic actions are unveiled. Similarly to other unsupervised approaches that rely on similar assumptions \cite{KuklevaKSG19}, our focus is on representation learning. However, similarly to \cite{SarfrazM0DGS21, 9879786}, our method takes as input a single video and doesn't care whether the same action is present in similar videos or not. This has considerable practical advantages for downstream applications since it can be in principle applied to any video no matter the dataset it belongs to nor if there exist videos having a similar temporal structure.
%Specifically, we learn the mapping function, parameterized by a shallow neural network, by independently minimizing an objective function for each sequence. The loss effectively models temporal smoothness and semantic similarity through an efficient tuplet selection strategy. 
%The main contribution of this work is to introduce a self-supervised training methodology that exploits both temporal and semantic relatedness in a very intuitively manner, resulting in temporal-semantic aware (TSA) representations suitable for the video action segmentation task. 
%Our technical contribution is a novel approach to action representation learning that uses a shallow network and a triplet loss operating on similarity distributions with a novel triplet selection strategy based on a downsampled temporal-semantic similarity weighting matrix. Our approach outperforms the state-of-the-art in action segmentation on \textit{Breakfast} and \textit{Youtube INRIA Instructional} benchmark datasets.
Action segmentation is obtained by applying a generic clustering algorithm on the learned temporal-semantic aware (TSA) representations (see Figure \ref{fig:tsne}). 
Our contributions are as follows: 

\begin{itemize}
    \vspace{-0.75em}
    \item We introduce the  novel approach to action representation learning that uses a shallow network and the single video itself as input, without the need for additional training data.
    \vspace{-0.7em}
    \item We demonstrate the effectiveness of using a triplet loss operating on similarity distributions with a novel triplet selection strategy based on a downsampled temporal-semantic similarity weighting matrix for the task of action segmentation.
    \vspace{-0.7em}
    \item We detail ablation study and we achieve state-of-the-art metrics on the \textit{Breakfast} and \textit{Youtube INRIA Instructional} benchmark datasets without using any training data than a single video itself as input.

\end{itemize} 

%-------------------------------------------------------------------------
%\vspace{-1.5em}
\section{Related work}
    \vspace{-0.5em}
\label{sec:related_work}
\begin{figure*}[t]
\centering
\includegraphics[width=0.95\linewidth]{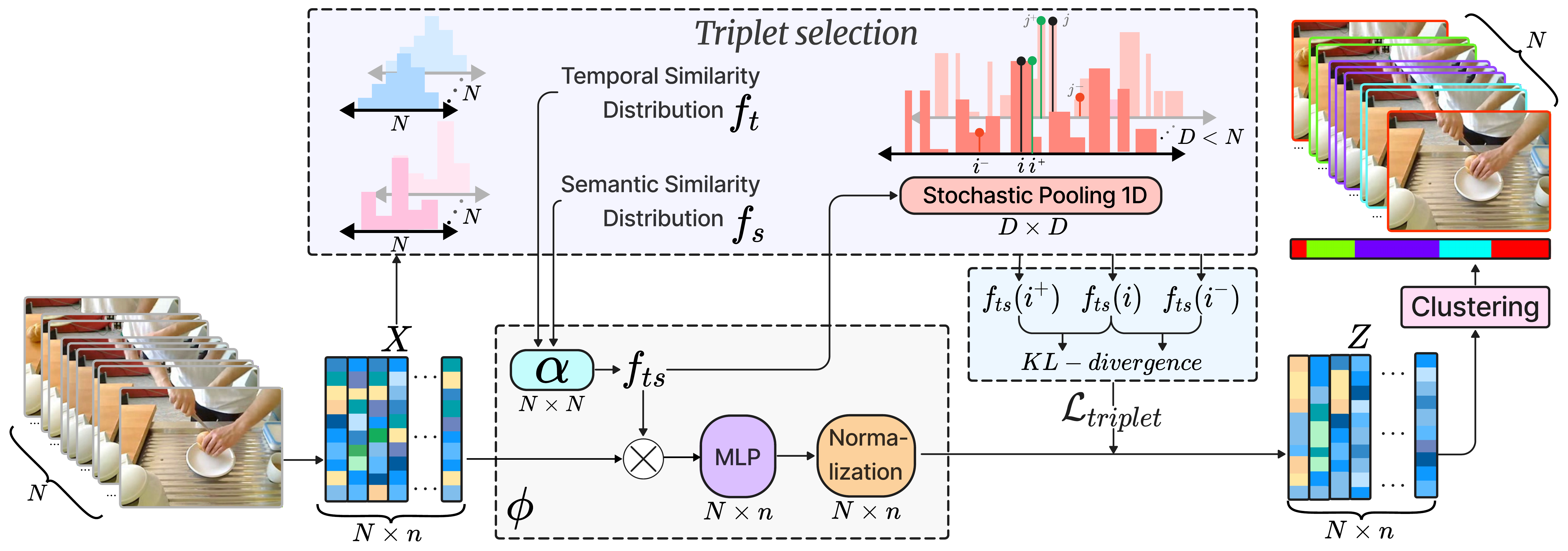}
\vspace{-1.2em}
\caption{Overview of the proposed TSA framework illustrated on a sample video of the Breakfast Dataset: network architecture transforming the initial features $X$ into the learned features $Z$ through a shallow network with a novel triplet selection strategy and a triplet loss based on similarity distributions.}
 \label{fig: overview}
\vspace{-1em}
\end{figure*}
\noindent\textbf{Fully supervised approaches.}
Action Segmentation has been traditionally tackled as a supervised learning problem. Existing approaches belonging to this category differ mainly in the way temporal information is taken into account by the model. 
Traditional approaches follow a two-step pipeline, that first generates frame-wise probabilities and then feeds them to high-level temporal models as in the Hidden Markov Model Tool Kit (HTK) approach \cite{KuehneGS16} or in \cite{NgHVVMT15}, which is based on recurrent neural networks. Lately, there has been a proliferation of models based on temporal convolutions to directly classify the video frames. Encoder-Decoder Temporal Convolutional Networks (ED-TCNs) \cite{LeaFVRH17} use a hierarchy of temporal convolutions to perform fine-grained action segmentation, but they can act solely on low-temporal resolution videos.  Instead, Multi-Stage Temporal Convolutional Network (MS-TCN and its improved version MS-TCN++) can act on the full temporal resolution of the videos and achieves increased performance \cite{FarhaG19, li2020ms}. Spatio-temporal convolutional layers \cite{LeaRVH16} have shown promising results in capturing temporal dependencies while being easier to train than previous methods.

The main drawback of traditional supervised approaches to action segmentation is the requirement of a large amount of quality labelled data for training, which limits their applicability to large-scale domains outside of existing pre-segmented datasets \cite{KuehneGS16, li2020ms}. 

\noindent\textbf{Weakly and semi-supervised approaches.}
To alleviate the need for large annotated datasets, weakly supervised techniques for video segmentation involve using transcripts (ordered list of the actions occurring in the video), visual similarities, and audio information to generate pseudo-labels for training \cite{FayyazG20}.  
%Weakly supervised video segmentation techniques have been developed to reduce the need for large annotated datasets. These approaches typically use transcripts or other forms of human-generated information to generate pseudo-labels for training. 
In \cite{KuehneRG17}, a Gaussian Mixture Models + Convolutional Neural Networks (GMM+CNN) is first initialized and used to infer the segments of a video given a transcription of it. The new segmentation is used to re-estimate and update the model parameters until convergence. In \cite{RichardKG17}, a recurrent neural network is used to model a discriminative representation of subactions, and a coarse probabilistic model to allow for temporal alignment and inference over long sequences.
Some approaches use machine learning models to infer the segments of the video \cite{Lu_2021_ICCV}. Other approaches, such as those based on frame-to-frame visual similarities \cite{HuangFN16}, self-attentions mechanism \cite{NG2021103298} or iterative soft boundary assignment \cite{DingX18}, enforce consistency between the video and labels without the need for temporal supervision. In the work \cite{FayyazG20}, a network is trained on long videos which are only annotated by the set of present actions and are trained by dividing the videos into temporal regions that contain only one action class and are consistent with the set of annotated actions. The work in \cite{RichardKIG18} adopts a Hidden Markov Model grounded on a Gated Recurrent Unit (GRU) for labelling video frames. This model has been subsequently improved in \cite{LiLT19-8}, where the model is trained through a Constrained Discriminative Forward Loss (CDFL) that  accounts for all candidate segmentations of a training video, instead than a single one. \cite{AlayracBASLL16} uses speech as additional sources of human-generated information in a weakly-supervised framework. 

Recent work has proposed a semi-supervised approach \cite{SinghaniaRY22} (ICC) consisting of a previous unsupervised training with a contrastive loss followed by a supervised training step with a small amount of labelled samples. These methods are limited to videos with transcripts and cannot be generalized to unconstrained videos.

\noindent\textbf{Unsupervised learning approaches.}
Unsupervised learning approaches typically learn action representation in a self-supervised fashion and then apply a clustering algorithm to obtain the action segmentation (assuming that the number of clusters is known).
Some methods model that minimizes prediction errors exploiting the order of scripted activities \cite{KuklevaKSG19} or combining temporal embedding with visual encoder-decoder pipelines \cite{VidalMataSKCK21}. Other approaches use deep learning architectures, including an ensemble of autoencoders, and classification networks that exploit the relation between actions and activities \cite{Li2021ActionSA}. Based on the assumption that in task-oriented videos actions occur in a similar temporal space, \cite{KuklevaKSG19, SenerY18} learn a strong temporal regularization that partially hides visual similarity. \cite{WangCLLXTF22} propose a Self-Supervised Co-occurrence Action Parsing method (SSCAP), for unsupervised temporal action segmentation which takes the recurrence of sub-actions into account in estimating the temporal path, and hence is able to handle complex structures of activities. Recently, \cite{Kumar_2022_CVPR} proposed a joint self-supervised representation learning and online clustering approach, which uses video frame clustering as a pretext task and hence directly optimizes for unsupervised activity segmentation (TOT+TCL).
   
Even if these approaches do not require labelled data, they are data-hungry and are not suitable for transferring the learned knowledge to a dataset with a different distribution and they demonstrated modest performance for the task of action segmentation at the video level. In contrast, we aim at unveiling the clusters underlying a single video. 
There is limited work in the literature on learning action representations in a self-supervised manner within a single video. \cite{AakurS19} proposes a model based on an encoder LSTM architecture with Adaptive Learning (LSTM+AL) that minimizes the prediction error of future frames and assigns segmentation boundaries based on the prediction error of the next frame. Some recent works proposed to learn event representations \cite{DiasD18-0, Dimiccoli2019EnhancingTS} and the underlying graph structure from a single sequence (DGE) \cite{DimiccoliW21}, where temporal windows are taken into account instead of all frames. %, but these have only been tested on low-temporal resolution image sequences.
In both cases, these approaches were not tested for the action segmentation of complex activity videos (high-temporal resolution), but only on low-temporal resolution image sequences. %, which involves different challenges. 

\noindent\textbf{Fully unsupervised approaches.}
Clustering methods, which generate a partition of the input data based on a specific similarity metric, have been poorly investigated within the field of action segmentation. However, very recent work \cite{SarfrazM0DGS21} has shown that simple clustering approaches, i.e. $K$-means, are instead a strong baseline for action segmentation. They hence proposed a new clustering approach called Temporally-Weighted FINCH (TW-FINCH), which is similar in spirit to the clustering approach named FINCH \cite{SarfrazSS19} but takes into account temporal proximity in addition to semantic similarity.
%Here we propose to learn temporally and semantically aware representations (TSA) before applying a generic clustering algorithm. 
Recently, \cite{9879786} proposed to detect action boundaries (ABD) by measuring the similarity between adjacent frames based on the insight that actions have internal consistency within and external discrepancy across actions. We based our approach on the same insight that we modelled via a deep metric learning approach.

%-------------------------------------------------------------------------
\section{Methodology}
\label{sec:methodology}
We assume that the representational clustering grounding action segmentation encodes both temporal and semantic similarity, based on two observations: (\textit{i}) an action in a video is a sequence of images temporally continuous, therefore temporal adjacent frames are likely to belong to the same action. (\textit{ii}) frames corresponding to the same action (but not necessarily temporal adjacent) should have similar representation, encoding the common underlying semantic. 

Formally, let $X\in{R}^{N\times n}$ denote the matrix of $n$-dimensional feature vectors for a given sequence of $N$ frames. We aim at learning a parametric function $\phi$ such that given the input feature matrix $X$, new Temporal-Semantic Aware (TSA) (see Figure \ref{fig: overview}) representations $Z \in \mathbb{R}^{N \times n}$ are obtained as $Z = \phi(X).$

\begin{figure*}[t]
  \centering
  \begin{subfigure}{0.9\linewidth}
  \centering
    %\fbox{\rule{0pt}{2in} \rule{.9\linewidth}{0pt}}
    \includegraphics[width=0.9\linewidth]{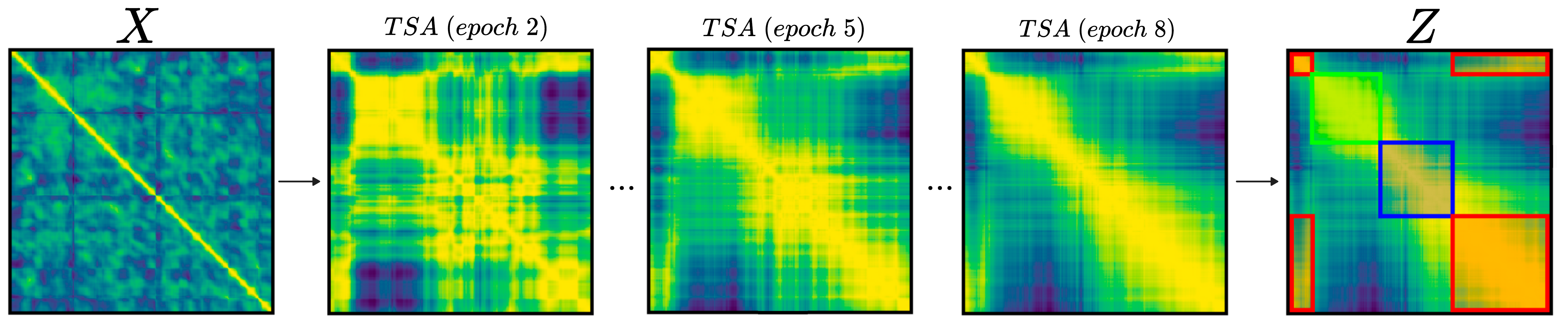}
    \vspace{-0.5em}
    \caption{}
    \label{fig:a}
  \end{subfigure}
  \hfill
  \begin{subfigure}{0.75\linewidth}
  \centering
    %\fbox{\rule{0pt}{2in} \rule{.9\linewidth}{0pt}}
    \includegraphics[width=1.0\linewidth]{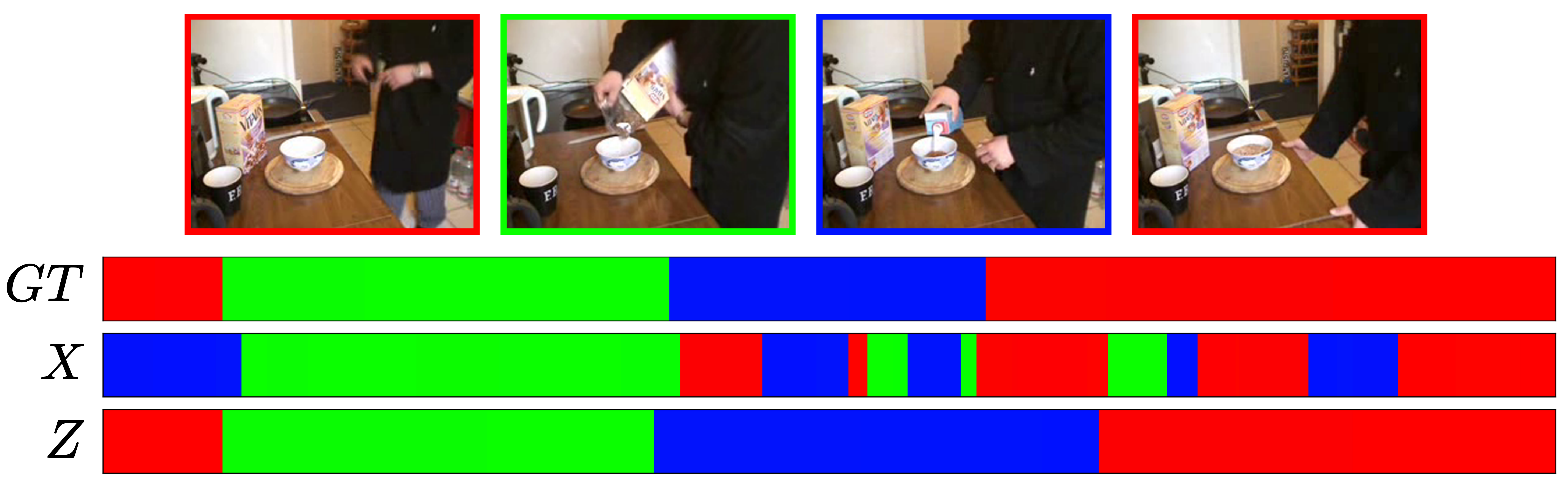}
    \vspace{-1.5em}
    \caption{}
    \vspace{-0.5em}
    \label{fig:b}
  \end{subfigure}
  \caption{Example of training and result obtained by using the TSA approach for a sample video of the \textit{Breakfast Action Dataset} (P34\_cereals). \textbf{(a)} Cosine similarity affinity matrix for initial features $X$ and evolution of the learned features $Z$ for different training epochs. Actions are highlighted as neighbour communities referring to the segmentation of the video when a clustering algorithm is applied. \textbf{(b)} Segmentation plots showing the ground truth, and the result of applying the same clustering algorithm to initial features $X$ (IDT) and  to the learned features $Z$ (TSA).}
  \label{fig:similarity_features}
  \vspace{-1.0em}    
\end{figure*}   

\vspace{-0.5em}   
\subsection{Triplet loss and triplet selection} To learn $\phi$, we minimize a triplet loss function (defined in Equation \ref{eq:loss}) that implements an original approach to select the triplets appropriately by relying on temporal-semantic similarity distributions $f_{ts}$ obtained as the weighted sum of the temporal and the semantic similarity distributions, say $f_t$ and $f_s$,  $f_{ts}= \pmb{\alpha} \cdot f_t + (1-\pmb{\alpha})\cdot f_s,$ where $\pmb{\alpha} \in [0,1]^{N\times 1}$ a vector of learning parameters of the function $\phi$. Under the model assumptions, it is easy to see that the similarity of $f_{ts}(k)$ and $f_{ts}(k')$ will be large when $k$ and $k'$ belong to the same event, and it will be small when $k$ and $k'$ belong two different events.
\vspace{-0.8em}
\paragraph{Semantic similarity distribution:}
To define $f_\mathcal{S}$, we assume that the set of most similar frames in the original feature space of an anchor $i$ is very likely to be part of the same action. The similarity of an anchor $i$ to all other frames is defined element-wise via a pairwise similarity, upon normalization to the total unit weight, $f_{\mathcal{S}} = w_{ij}/W,$ with $W= \sum_{i,j)\in E} w_{ij} $ and $w_{ij} = exp(-(1-d(x_{i}, x_{j}))/h)$, and where $E$ is the set of pairwise relations, $d(\cdot,\cdot)$ is the cosine distance and $h$ is the filtering parameter of the exponential function. The resulting pairwise similarities are normalized to represent empirical joint probability distributions between pairs of elements in the sequence. 
\vspace{-0.8em}
\paragraph{Temporal similarity distribution:}
To define $f_t$, we assume that as we move away from the anchor $i$, the likelihood of a feature vector $x_{j\neq i}$ to represent the same action as frame $i$ decreases. To model this behaviour, we define a weight function $w(\cdot)$ that depends on the temporal frame distance $d$ from the given frame as $w(d) = -1 + 2 \exp (-\frac{1}{\beta}d)$ where $\beta$ is a constant that controls the slope of the weight function and  $d$ is the temporal distance between frames. By imposing that $w(L/2)=0$, and then solving for $\beta$, we get that the constant $\beta$ can be expressed in terms of the positive window length, that is: $\beta = -L/(2 \ln (\frac{1}{2}))$. 
%we define a weight function to be $0$ at the boundary of $L$, thus giving a positive weight to samples temporally closer to the anchor and a negative weight to samples temporally far away from the anchor. By imposing the such condition, say $w(L/2)=0$, and then solving for $\beta$, we get that the constant $\beta$ can be expressed in terms of the positive window length, that is: $\beta = -L/(2 \ln (\frac{1}{2}))$. 

The temporal and semantic distributions are downsampled to reduce computational costs using stochastic pooling during the training. An anchor index is randomly selected from the set of downsampled indices $i\in D$. Its set of positive samples $\mathcal{P}_{i}$ is taken as the  $5\%$ of the frames with the highest similarity values in $i$-row of the temporal-semantic affinity matrix $f_{ts}$. %In (deep) metric learning, using hard negatives is an effective way to improve training speed. Our goal is to design a strategy for selecting hard negatives in an unsupervised setting that doesn't add extra computational overhead. To achieve this, 
We define the negative set $\mathcal{N}_{i}$ as the frames whose $i$-row $f_{ts}$ is between the mean and the sum of the mean and standard deviation of the similarity metric.
%from self-attention weights and a Temporal-Semantic similarity distribution $\mathcal{\rho_{TS}}$. This is represented by two similarity distributions: the temporal similarity distribution $\mathcal{\rho_{T}}$ and the semantic similarity distribution $\mathcal{\rho_{S}}$. These matrices measure how similar the points are in terms of time and meaning, respectively, and help ensure that $Z$ accurately captures the relationships between the points in $X$. We propose a simple but effective strategy that relies on the use of a temporal-semantic similarity distribution obtained as the weighted sum of the temporal and the semantic similarity distributions, $\rho_\mathcal{TS}= \alpha\cdot \rho_\mathcal{T} + (1-\alpha)\cdot \rho_\mathcal{S},$ where $\alpha \in [0,1]^{N\times 1}$ a vector of learning parameters of the function $\phi$. 
Our triplet loss is defined as:  
\begin{multline}
    \label{eq:loss}
   \mathcal{L}_{triplet}= \frac{1}{D}\sum_{i\in D}\max(0, \,KL(f_{ts}(i)||f_{ts}(i^{-})) \\
                                    - KL(f_{ts}(i)||f_{ts}(i^{+}))  
\end{multline}
where $KL$ represent the KL-divergence of the temporal-semantic similarity distribution $f_{ts}$. For each loss term, given an anchor index $i\in D$ with $D<N$, we define the triplet $\{i,\,i^{+},\,i^{-}\}$ where $i^{+}\in \mathcal{P}_{i}$ and $i^{-}\in \mathcal{N}_{i}$ that they are the sets of positive and negative indices, respectively.
%\vspace{-0.8em}
 
Using probability distribution functions (PDFs), $f_{ts}$ as feature vectors, instead of initial features $X$ can provide several benefits for action segmentation. Since PDFs consider all the information in the video for each frame, they are smoother and more robust compared to the initial features extracted from the data $X$ (they can be noisy or contain irrelevant information). % Moreover, probability distribution functions can compress information from the original feature space $X$, help identify relevant features, can be adapted to different tasks and domains, and provide a measure of uncertainty in the action segmentation results.
In the ablation study (see Section \ref{sec:4.3}), we will prove that this can improve the accuracy and robustness of the triplet loss for action segmentation.% and provide additional information that can be useful for a wide range of applications. 

%This similarity measure is very useful mainly in temporal segmentation, where frames that are similar to each other are grouped together as belonging to the same event \cite{DimiccoliW21, Dimiccoli2019EnhancingTS}}
%-------------------------------------------------------------------------

\section{Experimental evaluation}
\subsection{Experimental setup}

\paragraph{Datasets:} We report results on two widely used temporal action segmentation datasets: (1) the \textit{Breakfast Action Dataset}  \cite{6909500} % is one of the largest fully annotated datasets available for temporal and action segmentation. It 
consists of $10$ activities related to breakfast preparation, performed by $52$ different individuals in $18$ different kitchens for a total of $1712$ individual videos of $64$ dimensions each feature vector; (2) the \textit{Youtube INRIA Instructional Dataset} \cite{AlayracBASLL16} consists on $150$ instructional videos from YouTube whose feature vectors are $3000$-dimensional. These include $5$ different unrelated activities (changing tire, preparing coffee, performing a Cardio Pulmonary Resuscitation, jumping a car and repotting a plant), lasting 2 minutes on average. The most challenging part of this dataset is the amount of background frames present in each video, which reaches up to 63.5\% of the frames in the dataset. %and, following a standardized practice for evaluation, a fraction of those are discarded when validating the model. %The \textit{50 Salads Dataset} \cite{50Salads} comprises videos capturing $25$ individuals preparing $2$ mixed salads each, for a total of $50$ annotated videos. This dataset has two set of annotations, that reflects different granularities of the segmentation (coarse, detailed).
\vspace{-1.5em}
\paragraph{Evaluation metrics:} We use a clustering algorithm on the learned representations to segment the video into its atomic actions. To match the predicted segmentation labels and the ground truth, we follow the Hungarian matching algorithm to obtain a one-to-one mapping between the predicted labels (cluster index) and the ground truth at video-level.
Following previous work \cite{SarfrazM0DGS21, AakurS19, SenerY18, 9879786}, we report three widely used metrics:  (\textit{i}) accuracy of the segmentation and action identification, computed as the \textit{Mean over Frames (MoF)} metric, which indicates the percentage of action predicted frames correctly labelled after the Hungarian matching. (\textit{ii}) Similarity and diversity of the predicted segments, calculated as the \textit{Intersection over Union (IoU)} metric, also known as the Jaccard Index or Jaccard similarity coefficient. (\textit{iii}) The \textit{F1-score} computed across the predicted segments and the known ground truth to evaluate the quality of the action segmentation. Note that, among unsupervised methods, \cite{SenerY18, KuklevaKSG19} look for a one-to-one global matching (shared across all videos of a given task) instead that at the video-level. This generally leads to poorer performance than computing the the Hungarian at video-level.
\label{sec:experiments}

\subsection{Implementation details}
\paragraph{Input features.} To ensure a fair comparison to state-of-the-art methods targeting the action segmentation task, we use the same datasets and input features for the frame-level initial representations as in \cite{AakurS19, SenerY18, VidalMataSKCK21, SarfrazM0DGS21, 9879786, KuklevaKSG19}. For the \textit{Breakfast Action} we use the Improved Dense-Trajectory (IDT) \cite{WangS13a} features. These were provided by the authors of CTE \cite{KuklevaKSG19} in their open-sourced implementation. For \textit{Youtube Inria Instructional} we use a set of frame-level representations given by their original authors, which are features vectors formed by a concatenation of HOF descriptors and features extracted from VGG16-\textit{conv5} network.
\vspace{-1.5em}
\paragraph{Model architecture}
We used a shallow neural network consisting in our case of a multi-layer perceptron with a single hidden layer followed by a ReLu activation function. This makes our approach (Figura \ref{fig: overview}) easy to train and more suited for practical applications than existing approaches consisting of multiple convolutional layers and/or recurrent networks. Empirical experiments showed that using a single hidden layer was easier and faster to train than deeper models while achieving similar performance. The reported results are also invariant to the number of units in the hidden layer. 
\vspace{-1.5em}
\paragraph{Model training.}  The parameter $L$ used in this paper is the average number of action classes for a specific dataset, 6 and 9 for BF and YII, respectively. 
Empirical experiments showed that using a single hidden layer was easier and faster to train than deeper models while achieving similar performance. The reported results are also invariant to the number of units in the hidden layer. The architecture used to obtain our features is a multi-layer perceptron with as many units as the input feature dimensionality, $n$, although this could  be changed to obtain the desired output dimensionality. The batch size is equivalent to the downsampling and the number of batches will be the quotient of the number of frames and the batch size. We define the distance hyperparameter as the minimum threshold $\varepsilon$ that the difference of the last two losses should take. This hyperparameter is set to track early stops with a patience of 2 times. The minimum and maximum training epochs are fixed at 2 and 50, respectively. The initial learning rate depends on each dataset and follows an exponential learning decay rate of $0.3$ and a weight decay $L_2$ of $10^{-3}$ as the regularisation parameter.

\subsection{Model study} 
\label{sec:4.3}
Figure \ref{fig:similarity_features} illustrates how the initial features are modified during training to gradually unveil the representational clusters. On the left, we visualize the affinity matrix computed from the feature matrix on the original feature space. In the middle, we visualize the affinity matrix computed at different epochs during training and finally, on the right, our learned representations at the end of the training. Finally, the clusters, that were completely hidden on the initial affinity matrix, become more and more visible along the diagonal. The same label appears at different time intervals in the off-diagonal clusters, as reflected by the ground truth.
\vspace{-1.5em}
\paragraph{Ablation study.}
%To obtain the final segmentation, three different clustering algorithms were applied: K-means, Spectral clustering, and FINCH \cite{SarfrazSS19}. 
In Table \ref{tab:ablation_study}, we show the importance of modelling both temporal and semantic similarities, by using both $f_t$ and $f_s$ (being $f_{ts}$ when both are marked) and the effectiveness of our network as opposed to adding one more hidden layer or using another function, Radial Basis Function Neural Networks (RBFNN). We use it to compare with our problem because RBFNN are particularly good at modelling non-linear decision boundaries, making RBF networks well suited to our problems \cite{asvadi2011efficient}. In Table \ref{tab:comparision_pds}, we can see that by representing a frame through a PDF instead of a simpler feature vector, we can significantly improve the accuracy of action segmentation. Without this representation, the approach may not accurately capture the similarity between points in $X$, leading to segmentation errors. In addition, the average calculation time tells us that this improvement is achieved with no additional cost.
\begin{table*}
    \centering
    \begin{subtable}{0.5\textwidth}    
    
    %\noindent\setlength{\tabcolsep}{1pt}
    \resizebox{0.9\textwidth}{!}{ 
    \begin{tabular}{p{0.7\linewidth} |
                   p{0.1\linewidth}  
                   p{0.1\linewidth}
                   c |
                   c} 
            \toprule
            \multicolumn{5}{c}{\textbf{Breakfast Action Dataset}} \\
            \toprule
            \textbf{Baselines}  &  \textbf{MoF} & \textbf{IoU} & \textbf{F1} & \textbf{T} \\
            \hline 
            Equal Split* &  34.8 & 21.9 &  - &   \xmark \\
            Spectral*  & 55.5 & 44.6   & - &  \xmark \\
            Kmeans*  & 42.7 & 23.5 &  - &  \xmark  \\ 
            FINCH* \cite{SarfrazSS19} & 51.9 & 28.3 &  - &  \xmark \\             
            \hline
            \textbf{Unsupervised  } & & &  & \\
            \hline 
            LSTM+AL \cite{AakurS19} & 42.9 & 46.9 & - & \cmark \\
            VTE \cite{VidalMataSKCK21}  & 52.2 & - & - & \cmark \\            
            DGE*\cite{DimiccoliW21} \small{(Kmeans)}  & 58.8 & 47.8 & 51.6 &\xmark   \\
            DGE*\cite{DimiccoliW21} \small{(Spectral)}  & 59.5 & 48.5 & 51.7 &  \xmark   \\
            TW-FINCH \cite{SarfrazM0DGS21}  & 62.7 & 42.3 & 49.8 & \xmark   \\
            ABD \cite{9879786}  & \underline{64.0} & - & 52.3 & \xmark   \\
            \textbf{Ours*} \small{(Kmeans)}  & 63.7 & \textbf{53.3} & \textbf{58.0} & \xmark   \\
            \textbf{Ours*} \small{(Spectral)}  & 63.2 & \underline{52.7} &  \underline{57.8} &  \xmark  \\
            \textbf{Ours*} \small{(FINCH)} & \textbf{65.1} & 52.1 & 54.6 &  \xmark  \\       
            \hline
        \end{tabular} }
    \end{subtable}%   
    \begin{subtable}{ 0.5\textwidth}   
    \centering
    %\noindent\setlength{\tabcolsep}{1pt}
    \resizebox{0.9\textwidth}{!}{ 
    \begin{tabular}{p{0.70\linewidth} |
                   p{0.1\linewidth} 
                   p{0.1\linewidth} |
                   c} 
        \toprule
        \multicolumn{4}{c}{\textbf{Youtube INRIA Instructional Dataset}} \\
        \toprule
    \textbf{Baselines} & \textbf{F1} &  \textbf{MoF}   & \textbf{T} \\  [0.3ex]
    \hline 
    Equal Split*  & 27.8 & 30.2  & \xmark\\
    Spectral* & 44.6 & 55.1   & \xmark\\
    K-means* & 29.4  & 38.5& \xmark \\ 
    FINCH* \cite{SarfrazSS19}  & 35.4 & 44.8  & \xmark \\
    \hline
    \textbf{Unsupervised} &   & &   \\ [0.3ex]
    \hline
    LSTM+AL \cite{AakurS19}  & 39.7 & -  & \cmark \\
    DGE*\cite{DimiccoliW21} \small{(Kmeans)}& 47.0 & 42.1 & \xmark \\
    DGE*\cite{DimiccoliW21} \small{(Spectral)} & 48.9 & 44.8  & \xmark  \\
    TW-FINCH \cite{SarfrazM0DGS21} & 48.2 & 56.7  & \xmark  \\
    ABD \cite{9879786}  & 49.2 & \textbf{67.2}  & \xmark   \\
    \textbf{Ours*} \small{(Kmeans)} & \textbf{55.3} & 59.7 & \xmark \\ %& \textbf{49.3}
    \textbf{Ours*} \small{(FINCH)} & \underline{54.7}  & \underline{62.4} & \xmark \\ %& \textbf{51.8}
    \hline 
    \end{tabular} } 
    \end{subtable}%
    \vspace{-0.5em}
    \caption{Action Segmentation results on the BF and YII dataset by applying the Hungarian matching at the video-level. T denotes whether the method has a training stage on target activity/videos. The dash indicates “not reported”. * denotes results computed by ourselves. The best and second-best results are marked in bold and underlined, respectively.}
    \vspace{-1.5em}
    \label{tab:breakfast_action} 
\end{table*} 
\begin{table}[t]
     %\noindent\setlength{\tabcolsep}{1pt}
    \resizebox{0.47\textwidth}{!}{ 
    \centering
    \begin{tabular}{cc|cc|cc|cc|cc} 
        \toprule
        & & \multicolumn{2}{c|}{\textbf{BF} (kmeans)} &\multicolumn{2}{c|}{\textbf{BF} (FINCH)} & \multicolumn{2}{c|}{\textbf{YII} (kmeans)}& \multicolumn{2}{c}{\textbf{YII} (FINCH)} \\ [0.5ex] 
        
    \textbf{$f_t$} &  \textbf{$f_s$}   & \textbf{F1}   & \textbf{MoF}& \textbf{F1}  & \textbf{MoF}& \textbf{F1} & \textbf{MoF}& \textbf{F1}& \textbf{MoF} \\  
    \hline 
    \xmark & \cmark & 38.6 & 44.4 & 35.3 & 49.0 & 46.8 & 52.8 & 44.1 & 53.7\\
    \cmark& \xmark& 57.7 & 63.5 & 54.0 &  64.6&  54.8 & 59.4 & 53.5 & 62.2\\
    \cmark& \cmark& \textbf{58.0}  & \textbf{63.7}& \textbf{54.6} & \textbf{65.1}&  \textbf{55.3} & \textbf{59.7} & \textbf{54.7}& \textbf{62.4}\\
    %\cmark& \cmark&58.1 &64.0&53.4&64.5&55.4 &60.0& \textbf{54.1}&\textbf{63.0}\\
    \hline 
    \vspace{-0.3em} \\
    \hline 
    \multicolumn{2}{l|}{\textbf{MLP 2 LAYER}} & 57.2  & 63.2 & 52.0 & 63.4 &  54.4 & 59.1 & 52.9 & 61.8\\ 
    \hline 
    \multicolumn{2}{l|}{\textbf{RBFNN}}  & 20.9  &  50.2 & 19.1 & 49.8 & 12.9 & 42.4 &  10.4 & 42.6 \\ 
    \hline
    \end{tabular}
    }    
    \vspace{-0.5em}
    \caption{Ablation studies the BF and YII datasets, showing the importance of modelling both temporal and semantic information, and the effectiveness of the single layer MLP network as opposed to adding one more layer or using RBFNN.}
    \vspace{-1.5em}
    \label{tab:ablation_study} 
\end{table}

\subsection{Experimental results}  
\begin{figure*}[t]
    \centering
     \begin{subfigure}{1.0\linewidth}
        %\vspace{100mm}
        \label{fig:ejem0}
        \centering
        \includegraphics[width=0.9\linewidth]{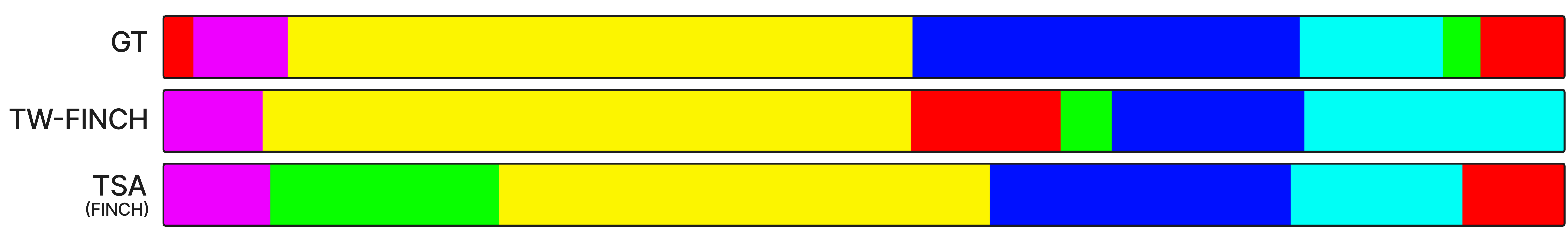}
        \subcaption{\textit{P48\_Sandwich,} BF - TSA $(72.6, 67.3)$ and TW-FINCH $(72.8, 50.5)$.}
    \end{subfigure}

   \begin{subfigure}{1.0\linewidth}
        %\vspace{100mm}
        \label{fig:ejem1}
        \centering
        \includegraphics[width=0.9\linewidth]{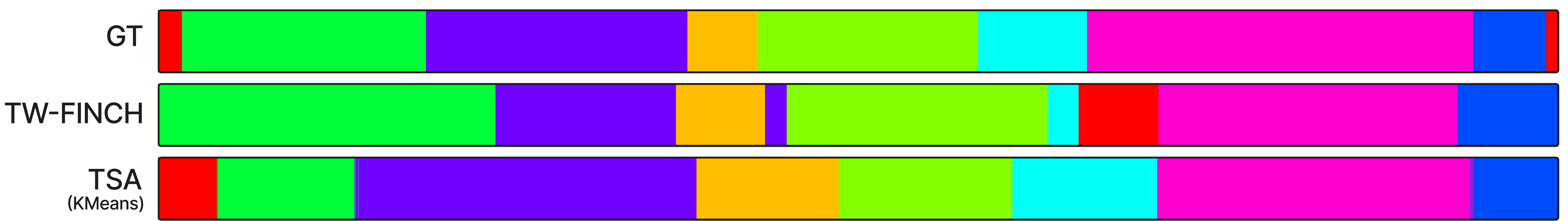}
        \subcaption{\textit{P45\_Scrambleegg,} BF - TSA $(77.7, 72.5)$ and TW-FINCH $(77.9, 68.1)$.}
    \end{subfigure}
   
    \begin{subfigure}{1.0\linewidth}
        \centering     
        \label{fig:ejem2}
        \includegraphics[width=0.9\linewidth]{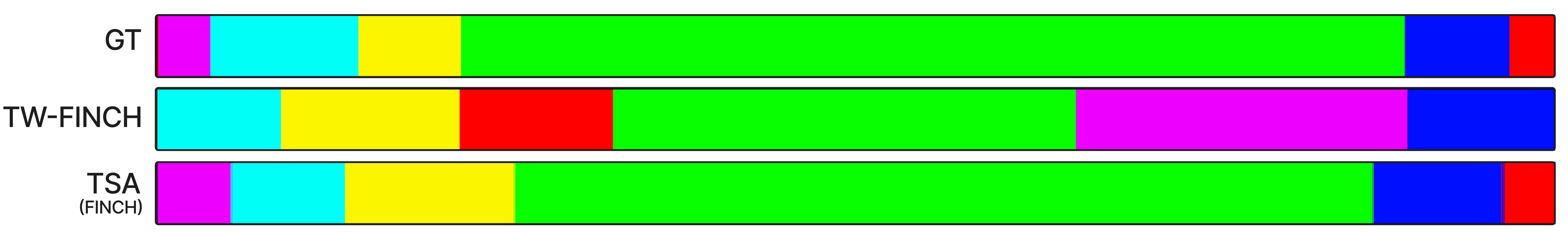}
        \subcaption{\textit{P18\_Friedeggs,} BF - TSA $(90.8, 85.6)$ and TW-FINCH $(73.7, 61.9)$.}
    \end{subfigure}
    
    \begin{subfigure}{1.0\linewidth}
        %\vspace{100mm}
        \label{fig:ejem3}
        \centering
        \includegraphics[width=0.9\linewidth]{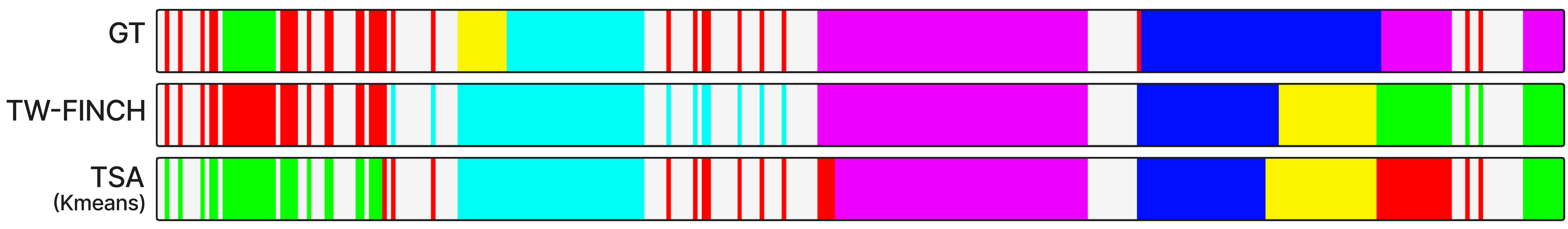}
        \subcaption{\textit{CPR\_0008, }YII - TSA $(62.5, 53.1)$ and TW-FINCH $(62.9, 48.4)$.}
    \end{subfigure}
    
    \begin{subfigure}{1.0\linewidth}
        %\vspace{100mm}
        \label{fig:ejem4}
        \centering
        \includegraphics[width=0.9\linewidth]{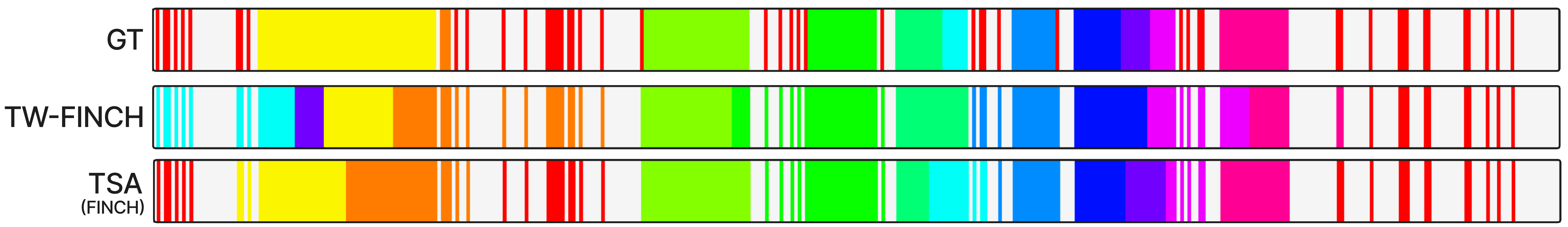}
        \subcaption{\textit{Changing\_Tire\_02, }YII - TSA $(76.1, 74.5)$ and TW-FINCH $(57.4, 53.2)$.}
    \end{subfigure}
    
    \begin{subfigure}{1.0\linewidth}
    
        %\vspace{100mm}
        \label{fig:ejem5}
        \centering
        \includegraphics[width=0.9\linewidth]{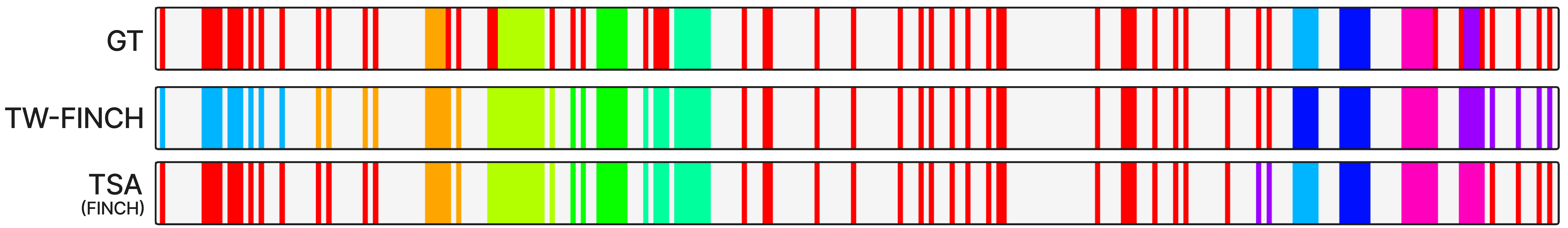}
        \subcaption{\textit{Jump\_Car\_0028,} YII - TSA $(65.7, 69.9)$ and TW-FINCH $(62.8, 64.2)$.}
    \end{subfigure}
    \vspace{-0.8em}
    \caption{Segmentation output comparisons on two sample videos from BF and YII. Each caption shows the name of the video and the results $(x,y)$ which are (MoF, F1) for each example.}
    \label{fig:segmentation_results}
    \vspace{-0.8em}
\end{figure*}

To obtain the final segmentation from the learned action representations we apply three different clustering algorithms: K-means, Spectral clustering and FINCH \cite{SarfrazSS19}. For comparison purposes, we report here the results of existing unsupervised methods that were computed by the proposing authors by applying the Hungarian matching at the video-level.  We used the code made publicly available by the authors \footnote{https://github.com/mdimiccoli/DGE} to compute the performance of DGE on the considered datasets, since this approach, similar to ours, computes a video representation suitable for the task of temporal/action segmentation.  
\vspace{-1.5em}
\paragraph{Breakfast Action:} The left-hand table \ref{tab:breakfast_action} reports the resulting metrics for the BF dataset, obtained with a learning rate $0.051$, distance $0.032$ and batch size $128$. Our method significantly outperforms all other existing approaches. Special attention on F1, which is considerably better in our results, which tells us better quality and less over-segmentation in our method. These results are consistent with all three clustering approaches considered for obtaining the final segmentation of our learned features. We can therefore conclude that TSA outperforms SoTA approaches for the downstream task of action segmentation. Examples of segmentation results on a few videos for this dataset can be seen in Figure \ref{fig:similarity_features} (b) and Figures \ref{fig:segmentation_results} (a)-(c). 

\begin{table}[t]
     %\noindent\setlength{\tabcolsep}{1pt}
    \resizebox{0.47\textwidth}{!}{ 
    \centering
    \begin{tabular}{c|c c c|c c c} 
        \toprule
        & \multicolumn{3}{c}{\textbf{BF} (FINCH)} &\multicolumn{3}{c}{\textbf{YII} (FINCH)} \\ [0.5ex] 
        
    \textbf{PDF}   & \textbf{F1} &  \textbf{MoF} &\textbf{Time} (seconds)  & \textbf{F1}& \textbf{MoF} &\textbf{Time} (seconds)\\  
    \hline     
     \xmark& 50.5 &   57.7 & 32.521 & 48.4 &  56.4 & 11.307 \\
     \cmark&    \textbf{54.6} &  \textbf{65.1}& 32.935 & \textbf{54.7}& \textbf{62.4} & 11.004 \\
     %\cmark& \cmark&58.1 &64.0&53.4&64.5&55.4 &60.0& \textbf{54.1}&\textbf{63.0}\\
     
    \hline 
    \end{tabular}}    
    %\vspace{-0.5em}
    \caption{Comparative results obtained by using the triplet loss with and without a PDF of similarities, and the average running time for a video in the BF (left) and YII (right) datasets.}
    \vspace{-1.0em}
    \label{tab:comparision_pds} 
\end{table}
\vspace{-0.8em}
 \paragraph{Youtube INRIA Intructional:} The right-hand table  \ref{tab:breakfast_action} reports the resulting metrics for the YII dataset, obtained with a learning rate $0.403$, distance $0.892$ and batch size $12$. This dataset is particularly challenging because of the nature of the annotations, where most of the frames in each video are labelled as background frames. To enable direct comparison, we follow the same procedure used in previous work \cite{SarfrazM0DGS21, 9879786, SenerY18} and report results  by removing the ratio ($\tau = 75\%$) of the background frames from the video sequence and then report the performance. To capture the temporal dependencies in the time window, we compute the temporal similarity matrix before subtracting a ratio of the background frames. Our method improves the best \textit{F1} metrics from the literature with a large margin, which indicates the quality of the segmentation in our method on both datasets, as the MoF does not reflect the quality, especially when the whole sequence is dominated by some very long segments. A segmentation output sample of our method for this dataset is plotted in Figures \ref{fig:segmentation_results} (d)-(f). %Our proposed approach with attention modules is beneficial under these circumstances. 
Also for this dataset, our results are consistent with all two clustering approaches.  
%The main challenge of this dataset when clustering the background of each video is that human-annotated background frames are semantically unrelated between them, thus the semantic term of our method does not capture a large part of the semantic relations between temporally non-adjacent background sets.

It is important to highlight the cases in Figures \ref{fig:segmentation_results} (a), (b) and (d) where the effectiveness of our approach can be clearly observed. In these examples, the MoF values are comparable in both clustering methods, with the TW-FINCH method exhibiting slightly higher MoF values. However, it is noteworthy that our F1 is significantly better which shows a better quality of segmentation.

%------------------------------------------------------------------------
\section{Conclusions}
\vspace{-0.8em}
\label{sec:conclusions}
This paper introduced a novel fully unsupervised approach for learning action representations in complex activity videos that solely operates on a single unlabelled input video. Our method exploits the temporal proximity and the semantic similarity in the initial feature space to discover the representational clustering grounding action segmentation.
Our key contributions are a shallow architecture and a triplet-based loss with a triplet-based selection mechanism based on similarity distribution probabilities to model temporal smoothness and semantic similarity within and across actions. Experimental results on the BF and the YII datasets demonstrated that the learned representations, followed by a generic clustering algorithm, achieve SoTA performance. Furthermore, it has the advantage of not requiring human annotations, is easy to train and does not present domain adaptation issues. 
Future work will consider how to jointly learn the action clusters and the representation as well as how to build on representations learn at the video-level to match videos at the activity-level.

\section*{Acknowledgments}
This work was supported by the project PID2019-110977GA-I00 funded by MCIN/ AEI/ 10.13039/ 501100011033 and by "ESF Investing in your future".
%\vfill
%\pagebreak
%%%%%%%%% REFERENCES
{\small
\bibliographystyle{ieee_fullname}
\bibliography{egbib}
}

\end{document}